# Multilingual Neural Machine Translation System for Indic to Indic Languages

Sudhansu Bala Das[1], Divyajyoti Panda[1], Tapas Kumar Mishra[1],
Bidyut Kr. Patra[2], Asif Ekbal[3]

[1*]National Institute of Technology(NIT), Rourkela, Odisha, India.
[2]Indian Institute of Technology (IIT), Varanasi, Uttar Pradesh, India.
[3]Indian Institute of Technology, Patna, India.
baladas.sudhansu@gmail.com, pandadivya02@gmail.com,
mishrat@nitrkl.ac.in, bidyut.cse@iitbhu.ac.in, asif@iitp.ac.in.

**Abstract**

Language is an effective means of communication that enables people to express their thoughts, feelings, desires, and other emotions. The method of automatically converting information from one language to another without human intervention is known as Machine Translation (MT). People are creating and using MT models to overcome these language barriers. Multilingual neural machine translation (MNMT), which builds a single model for multiple languages, is beneficial since it decreases the training time and improves low-resource translation. Low-resource regional languages, like Indian Languages (ILs), oversee the lack of good-quality MT models. In this paper, Indic-to-Indic (IL-IL) MNMT baseline models for 11 ILs are implemented on the Samanantar corpus and analyzed on the Flores-200 corpus. All the models are evaluated using the BLEU score. In addition, the languages are classified under 3 groups namely East Indo-Aryan (EI), Dravidian (DR), and West Indo-Aryan (WI). The effect of language relatedness on MNMT model efficiency is studied. Owing to the presence of large corpora from English (EN) to ILs, MNMT IL-IL models using EN as a pivot are also built and examined. To achieve this, English-Indic (EN-IL) models are also developed, with and without the usage of related languages. Results reveal that using related languages is beneficial for the WI group only, while it is detrimental for the EI group and shows an inconclusive effect on the DR group, but it is useful for EN-IL models. Thus, related language groups are used for the development of pivot MNMT models. Furthermore, the IL corpora are transliterated from the corresponding scripts to a modified ITRANS script, and the best MNMT models from the previous approaches are built on the transliterated corpus. It is observed that the usage of pivot models greatly improves MNMT baselines with AS-TA achieving the minimum BLEU score and PA-HI achieving the maximum score. Among languages, AS, ML, and TA achieve the lowest BLEU score whereas HI, PA, and GU perform the best. Transliteration also helps the models with few exceptions. The best increment of scores is observed in ML, TA, and BN and the worst average increment is observed in KN, HI, and PA, across all languages. The best model obtained is the PA-HI language pair trained on PA-WI transliterated corpus which gives 24.29 BLEU.

# 1 Introduction

Languages are essential for interactions among people, societies, and countries. They enable people to convey their thoughts, feelings, and ideas. Language facilitates communication, allowing people to create connections, exchange experiences, and define common goals. However, two people using different languages find it hard to communicate with each other without the use of a third party such as a translator. This is especially common in linguistically diverse countries like India, where people from various regions speak distinct regional languages. These regional languages help people to interact and transfer knowledge among themselves but also inhibit interaction across communities and regions. Due to the wide range of cultures, translation assistance is critical to facilitate communication among different parts of India. But, having human translators everywhere is sometimes not possible. Hence, a machine translation(MT) model is needed [1].

MT is a means of translating text from one language to another with the help of machines. Due to its speed, cost-effectiveness, and reliability, it has been increasingly used as a tool to assist in the accurate translation of texts instead of human translation. It is easier to train a machine to translate between any language pair than humans, making it more flexible and adaptable. Hence, researchers are developing appropriate approaches that do the job of translating from one language to another with as little human effort as possible [2]. However, MT models have to deal with many issues associated with language translation, such as ambiguity in meaning, cultural differences, and compound words. To mitigate these problems, different models have been developed over time.

For the past few years, there has been a significant increase in work on MT models involving more than two languages. Multilingual Neural Machine Translation (MNMT) models handle translations among multiple language pairs [3]. The ultimate objective of MNMT research is to create a single model for translation among several languages using as few linguistic resources as possible.

There are several reasons why multilingual models gained popularity. One of the advantages is that MNMT reduces the number of models to train and hence simplifies architecture and deployment, thus saving time and resources. It significantly reduces complexity as compared to multiple bilingual MT models, since only one model is to be trained instead of one between each pair of languages. It minimizes the number of model parameters, allowing for a more straightforward implementation. This makes it preferable over bilingual NMTs when more than 2 languages are involved. Secondly, the model can learn linguistic features from other languages present in the multilingual corpus which greatly increases the accuracy. It even allows translating among languages that were never coupled together in the training corpus [4] [5]. MNMT proves to be especially useful in the case of low-resource languages, many of which have not been worked upon adequately.

Hence, this paper illustrates the effectiveness of MNMT systems on low-resource language pairs. The objective of this work is to build baseline Indic-Indic (IL-IL) MNMT models to translate among 11 ILs, namely Assamese (AS), Bengali (BN), Gujarati (GU), Hindi (HI), Kannada (KN), Malayalam (ML), Marathi (MR), Odia (OR), Punjabi (PA), Tamil (TA), and Telugu (TE).

The languages used in the experiment are abbreviated in Table 1, which adheres to ISO 639-2 Code [6].

**Table 1**: Language codes

| Language | Code |
|---|---|
| Assamese | AS |
| Bengali | BN |
| English | EN |
| Gujarati | GU |
| Hindi | HI |
| Kannada | KN |
| Malayalam | ML |
| Marathi | MR |
| Odia | OR |
| Punjabi | PA |
| Tamil | TA |
| Telugu | TE |
| Thai | TH |

## 1.1 Motivation and Contribution

It is observed that much progress has been made in European languages since the development of MT, owing to their ubiquity and high resourcefulness, that is an abundance of corpora containing translation pairs [7], [8]. However, this leaves thousands of languages around the globe at a disadvantage, with not many good translation models available. The Indian subcontinent is a region of great ethnic and linguistic diversity, home to hundreds of languages across multiple language families. Some of them are spoken by millions of people across wide geographical areas and have sufficient corpora available, while many others are spoken by a couple of hundred people only in localized communities. In India, most people know and speak one or two ILs but find it difficult to understand other ILs. Due to the diversity, communication between people and societies speaking different languages becomes a challenge. This reflects the need to make better and improved translation models to convert texts from one IL to another. In light of the advantages of MNMTs in low-resource languages, this paper focuses to built an IL-IL MNMT model.

The following points outline the main contributions of this work:

1. This work presents the first attempt to investigate and build a baseline for an IL-IL MNMT model on the given 11 ILs.
2. The paper demonstrates whether language grouping (Dravidian(DR): KN, ML, TA, TE, East Indo-Aryan (EI): AS, BN, OR, and West Indo-Aryan (WI): GU, HI, MR, PA) benefits MNMT.
3. This work presents an analysis of Samanantar IL-IL corpus in terms of lexical diversity.
4. The use of EN as a pivot to enhance the translation quality of the models is explored.
5. It is the first attempt to study the effect of translated models on IL corpus transliterated to modified ITRANS, with mapping of EN characters to a foreign language (TH in this case).

## 2 Related Work

In this section, a few research works focusing on MNMT models have been shown.

In the year 2015, Dong et al. [9] have proposed a multitask learning method in one-to-many multilingual translation situations which uses one encoder for the source language and distinct attention methods and decoders with each target language. Their architecture can be used in instances with either large amounts of parallel or limited parallel corpus. Experiments reveal that their multi-task learning approach produces much superior translation quality compared to the individual-taught model in both cases on the publically accessible corpus.

Firat et al. [10] have presented a many-to-one (up to three languages) model of language-specific encoders and decoders and a single attention mechanism. They demonstrate effectiveness in low-resource settings and present a new finetuning method for the multiway MNMT, which allows for zero-resource MT. Despite the positive empirical results reported in their research, several flaws exist. For example, their experiments have only been conducted in three European languages: Spanish, French, and English. More analysis using various languages is required to reach a more definite result.

In 2017, Johnson et al. [4] worked on MNMTs where the sentence representations trained for various source languages tend to cluster based on the semantics of the source sentence instead of the language. Their models are additionally able to execute implicit bridging among language pairs that were never explicitly observed during training and demonstrate that neural translation is capable of zero-shot translation and transfer learning. Their suggested strategy has been verified to perform reliably in a Google-scale production setup, allowing them to grow to many languages quickly.

Schwenk and Douze et al. [11] train encoder-decoder frameworks on numerous source and target languages and explore the similarity of source sentence representations across languages. They present experimental proof that sentences that share comparable embedding spaces are linguistically connected, but possess distinct syntax and structure. They suggested a novel multilingual similarity-searching evaluation procedure easily scalable to numerous languages and massive corpora.

Aharoni et al. [3] conducted a significant experiment in a massively MNMT model training, translating as many as 102 languages from and to English within a single model. They investigate various training configurations and analyze the trade-offs between the accuracy of translation and numerous modeling decisions. Their findings are based on the publicly accessible TED Talks multilingual corpus, and they demonstrate that massively multilingual many-to-many models work efficiently in low-resource contexts, surpassing the prior state-of-the-art with enabling up to 59 languages.

In transfer learning environments, MNMT models have demonstrated significant empirical success. Kudugunta et al [12] used Singular Value Canonical Correlation Analysis (SVCCA), an underlying similarity structure, to evaluate illustrations throughout different languages, layers, and models, to comprehend massively MNMT representations (using 103 languages). They use separate encoder and decoder networks to model various language combinations in a many-to-many environment.

In the year 2019, Vazquez et al [13] approach the multilingual problem using a new technique. Language-specific encoders and decoders are proposed in their work but with a common independent attention mechanism. Their findings indicate that an attention bridge layer can efficiently exchange parameters within multilingual environments, improving by 4.4 BLEU points over baselines.

Angela Fan et al [14] develops a Many-to-Many MNMT model that is able to directly translate among any pair of 100 languages. They create and open-source a training data set with parallel data which encompasses thousands of language directions and was generated with large-scale mining. They investigate the way to successfully boost model capacity using a combination of dense scaling and language-specific sparse parameters for building high-quality models.

Recently, Eriguchi et al [15] demonstrated a way to practically develop MNMT systems that provide arbitrary X-Y translation directions while utilizing multilingual through a two-phase training approach involving pretraining as well as finetuning. Experimenting over the WMT'21 multilingual translation task, they show that their methods excel over the conventional baselines of direct bilingual models and pivot translation models in most directions, yielding +6.0 and +4.1 BLEU on average, without any architecture changes or additional corpus collection.

Aharoni et al [3] conducts a significant experiment in massively multilingual NMT model training, translating as many as 102 languages from and to English within a single model. They investigate various training configurations and analyze the trade-offs between the accuracy of translation and numerous modeling decisions. They present findings from a publicly accessible TED Talks multilingual corpus that demonstrate that massively multilingual many-to-many models perform well in environments with limited resources, surpassing the previous state-of-the-art while promoting up to 59 languages.

The paper is organized as follows. Section 3 describes the characteristics of Indian languages. The experimental setup and all methods are illustrated in section 4. In section 5 narrates the conclusion and future work.

## 3 Characteristics of Indic Languages

The Indian subcontinent is a region of vast linguistic diversity, home to 122 major languages spoken in India alone, according to the 2001 Indian census [16]. These languages are distributed across multiple language families (groups of families with similar vocabularies and basic grammatical units), including Indo-European, Dravidian, Sino-Tibetan, Austoasiatic, Austronesian, and Tai-Kadai language families [17]. Indo-European and Dravidian language families are considered the two major language families in India, with speakers spread over wide geographical regions [18]. While Indo-European languages are more concentrated in the North-Indian basins and southern islands, Dravidian languages are more concentrated in the peninsular regions in South India [19]. The corpus used in the experiment utilizes languages that come from these two categories.

### 3.1 Indo-European Language Family

Indo-European language family is a family of languages native to most of Europe, the Iranian plateau, and the Indian subcontinent [20]. It is considered the largest language family in the world, home to about 46 % of native speakers in the world. The languages belonging to this family are hypothesized to have originated from the Proto-IndoEuropean language. It hosts several widely used languages, such as English, Hindi, and Spanish. It can be categorized into many different branches, two of which used for the experiment are Germanic languages and Indo-Iranian languages [21].

### 3.1.1 Germanic languages

Germanic languages are languages largely native to German lands, Scandinavia, and the British Isles, but are now widespread across the world. It consists of widely spoken languages including English and Standard German[22]. These languages are said to have derived from Proto-Germanic and consist of some distinguishing characteristics, such as strong stress on the first vowel.

English is considered as the most widely spoken Germanic language [23]. It originated in the British Isles but is now the lingua franca in several places across the world. English is still used in most professional and academic settings. In linguistically diverse regions like India, English is considered the medium of communication between people who do not speak the same languages.

### 3.1.2 Indo-Iranian languages

Indo-Iranian languages are languages native to the Iranian plateau and North Indian plains. They are said to have originated from the Common Aryan language. Some widely spoken languages in this group include Hindi, Bengali, and Punjabi. It is divided into Indo-Aryan, Iranian, and Nuristani sub-branches [24].

Indo-Aryan languages are the languages that span across North India. A large number of ILs can be grouped under this category [25]. These languages are said to have originated from the proto-Indo-Aryan language which is very close to Vedic Sanskrit. They share a large degree of similarity in terms of vocabulary and grammatical structures. These languages use almost identical consonants and similar vowels, albeit with small differences. Indo-Aryan languages can be classified into Dardic, Northern, Northwestern, Western, Central, Eastern, Southern, and Insular Indo-Aryan languages [26]. Some common groups along with associated languages are described in Figure 1.

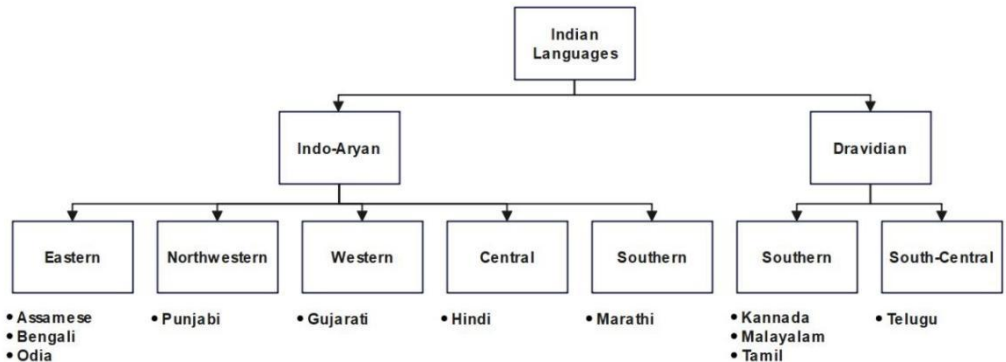

**Figure 1** Related Language Group

## 3.2 Dravidian Language Family

The Dravidian language family comprises of languages spoken across South India, north-east Sri Lanka, and north-west Pakistan [27]. Unlike Indo-Aryan languages, Dravidian languages are said to be indigenous to India and are said to be originated from proto-Dravidian, which might have been spoken in the Indus Valley civilization. The languages also share a great similarity in vocabulary and grammar, but they are notably distinct from the Indo-Aryan languages. They show several unique grammatical features like agglutination and clusivity, which are not found commonly in Indo-Aryan languages [28]. Here agglutination refers to the phenomenon where several morphemes (meaningless word units that contribute to the meaning of a word) can be combined in multiple ways to produce a whole set of new words and clusivity refers to the distinction made between inclusive "we" and exclusive "we", depending upon whether the second person pronoun is included or not respectively.

Kannada, Malayalam, Tamil, and Telugu are the four most popular Dravidian languages. Dravidian languages are divided into a few branches, namely South Dravidian, South-Central Dravidian, Central Dravidian, and North Dravidian [29]. Some common branches along with associated languages are described in Figure 1.

## 3.3 Indic scripts and the Roman Script

Indic scripts or Brahmi scripts are writing scripts that are used to write ILs, as well as languages used in South East Asia and parts of East Asia [30]. They are classified as abugidas, where each unit is a consonant letter with associated vowel markings, where a particular vowel (mid-central vowel/open-mid back rounded vowel) acts as an inherent vowel, implicit when the base character is used, whereas other vowels are designated using diacritics (signs which change the pronunciation of the letters which they are associated with). A special mark, called by several names like halant, hosonto or pollu, indicates the absence of the vowel.

On the contrary, English uses Roman script as the writing system [31]. Unlike Indic scripts, Roman Script is an alphabet, where each consonant and vowel is associated with a character of an equal stature [32]. English is notable for its use of lack of diacritics, where every letter may correspond to a class of sounds instead of a particular sound. Table 2 depicts which script is associated with which languages, along with examples.

**Table 2** Languages and associated scripts

| Scripts | Language | Example |
|---|---|---|
| Bengali | AS, BN | দ্রুত বাদামী শিয়াল অলস কুকুরের উপর ঝাঁপিয়ে পড়ে। |
| Roman | EN | The quick brown fox jumps over the lazy dog. |
| Gujarati | GU | |
| Devanagari | HI, MR | तेज, भूरी लोमड़ी आलसी कुत्ते के उपर कूद गई। |
| Kannada | KN | ತ್ವರಿತ ಕಂದು ನರಿ ಸೋಮಾರಿಯಾದ ನಾಯಿಯ ಮೇಲೆ ಜಿಗಿಯುತ್ತದೆ. |
| Malayalam | ML | പെടന്നുനുള്ള തവിടിക് കുറുക്ക മടിയനായ നായയുടെ മുകളിലൂടെ ചാടുന്നു. |
| Odia | OR | ଶୀଘ୍ର ବାଦାମୀ କୁମ୍ଭୀରର ଅଳସୁଆ କୁକୁର ଉପରେ ଡେଇଁ ପଡ଼େ। |
| Gurmukhi | PA | ਤੇਜ਼ ਭੂਰੀ ਲੂੰਬੜੀ ਆਲਸੀ ਕੁੱਤੇ ਦੇ ਉੱਪਰ ਛਾਲ ਮਾਰਦੀ ਹੈ। |
| Tamil | TA | விரைவான பழுப்பு நரி சோம்பேறி நாய் மீது குதிக்கிறது. |
| Telugu | TE | త్వరిత గోధుమ నక్క సోమరి కుక్కపై దూకుతుంది. |
| Thai | TH | สุนัขจิ้งจอกสีน้ำตาลกระโดดข้าม สุนัขขี้เกียจอย่างรวดเร็ว |

## 4 Experimental Setup

The workflow of the experiments is illustrated in Figure 2.

### 4.1 Dataset Analysis

For building IL-IL MNMT models, IL-IL Samanantar Corpus is utilized for training purposes [33]. It is one of the largest corpus available for ILs and contains millions of sentence pairs that are derived from two language families. Flores200 corpus is used for testing and validation purposes [34]. It has two corpora dev (997 lines) and devtest (1012 lines). Table 3 illustrates the statistics of the corpus. AS and OR languages are having the smallest corpora whereas TE and ML languages are having largest corpora in the IL-IL corpus. For the pivot models, English-Indic (EN-IL) Samanantar corpus is used which contains data from English to 11 ILs. From the table, it can be observed that AS and OR are having smallest corpora whereas HI and BN have the largest corpora in EN-IL language pairs.

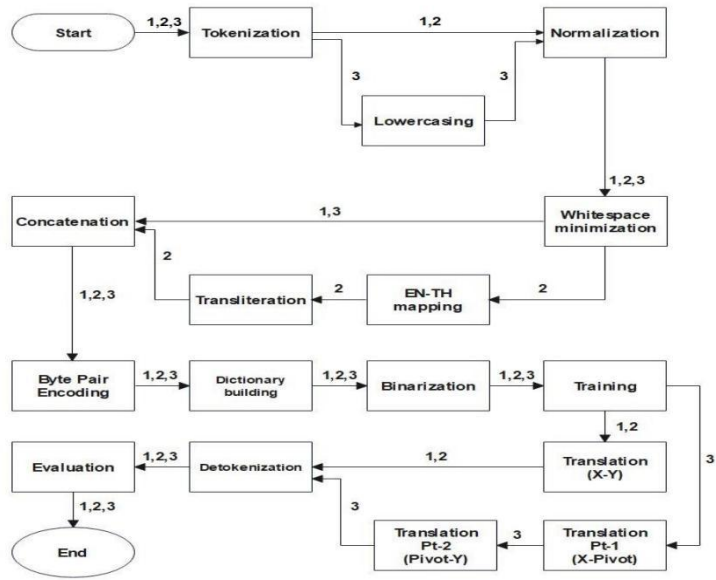

**Figure 2** Flowchart of Work( 1: MNMT Model, 2: MNMT and Transliteration, 3: Pivot MNMT)

**Table 3** Statistics of IL-IL and EN-IL corpora (in thousands)

|    | BN  | EN   | GU   | HI   | KN   | ML   | MR   | OR  | PA   | TA   | TE   |
|----|-----|------|------|------|------|------|------|-----|------|------|------|
| **AS** | 356 | 138  | 142  | 162  | 193  | 227  | 162  | 70  | 108  | 214  | 206  |
| **BN** |     | 8435 | 1576 | 2627 | 2137 | 2876 | 1876 | 592 | 1126 | 2432 | 2350 |
| **EN** |     |      | 3020 | 8444 | 4015 | 5780 | 3289 | 990 | 2401 | 5096 | 4776 |
| **GU** |     |      |      | 2465 | 2053 | 2349 | 1757 | 529 | 1135 | 2054 | 2302 |
| **HI** |     |      |      |      | 2148 | 2747 | 2086 | 659 | 1637 | 2501 | 2434 |
| **KN** |     |      |      |      |      | 2869 | 1819 | 533 | 1123 | 2498 | 2796 |
| **ML** |     |      |      |      |      |      | 1827 | 558 | 1122 | 2584 | 2671 |
| **MR** |     |      |      |      |      |      |      | 581 | 1076 | 2113 | 2225 |
| **OR** |     |      |      |      |      |      |      |     | 507  | 1076 | 1114 |
| **PA** |     |      |      |      |      |      |      |     |      | 1749 | 1756 |
| **TA** |     |      |      |      |      |      |      |     |      |      | 2599 |

## 4.2 Corrected Type/Token Ratio

Corrected Type/Token Ratio (CTTR) is a measure of lexical diversity of a text corpus [35] . It is defined by the following equation:

$$CTTR = \frac{\tau}{\sqrt{2.\rho}} \quad (1)$$

where, $\tau$ denotes number of unique tokens in the corpus, and $\rho$ denotes the total number of tokens in the corpus

A high CTTR score implies a greater ratio of unique tokens in the corpus, indicating a lexically rich language. Typically, ILs possess moderate to high CTTR because of their extensive use of grammatical cases and word genders. Dravidian languages like ML and TA possess even higher CTTR due to agglutination, where several morphemes (meaningless word units that contribute to the meaning of a word) can be combined in multiple ways to produce a whole set of new words. Without agglutination, these words would have been replaced by a set of words expressing the same meanings as the morphemes, which would have decreased the number of unique tokens in the corpus.

To analyze the CTTR of ILs, all files of Samanantar IL-IL and Flores200 corpora were segregated according to the associated languages and concatenated. Then, 1750000 lines were selected randomly (with replacement) from the concatenated file and CTTR is computed on the subset. This was done because CTTR is dependent on the size of the corpus. The statistics of the corpus are described in Table 4.

**Table 4** Corpus statistics (CTTR and number of lines)

| Language | CTTR | Lines (in thousands) |
|---|---|---|
| AS | 14.3 | 1842 |
| BN | 46.51 | 17924 |
| GU | 46.69 | 16366 |
| HI | 31.6 | 19470 |
| KN | 135.9 | 18173 |
| ML | 246.6 | 19835 |
| MR | 67.35 | 15497 |
| OR | 38.79 | 6221 |
| PA | 28.71 | 11340 |
| TA | 386.1 | 19823 |
| TE | 144.5 | 20460 |

The table shows that languages like HI and PA are low in lexical diversity, whereas agglutinative languages like ML and TA are higher in diversity.

## 4.3 Preprocessing

Different preprocessing steps used in the experiments are illustrated below.

### 4.3.1 Tokenization

Tokenization is the process of segmenting a sentence into tokens (words, numbers, and punctuations) [36]. It returns a list of tokens, or texts with consecutive tokens separated by a space. Its purpose is to allow the system to segregate and distinguish between different lexical units of a sentence. For instance, the sentence "Highways, airways, railways and waterways, all these have been integrated as per the requirement of each other." is transformed to "Highways , airways , railways, and waterways , all these have been integrated as per the requirement of each other .", where spaces have been added before commas and full stops.

There are several tokenizer models available. The Moses tokenizer [37] is considered as one of the standard models for tokenization, which effectively tokenizes several European languages including EN. However, it does not effectively ILs because it treats halants and nuktas (a special diacritic which transforms one consonant to its variant form) as separate tokens. Therefore, a corrected Moses tokenizer is used, which recombines these diacritics with surrounding words.

### 4.3.2 Normalization and Whitespace Minimization

Language is a fluid and complex structure with multiple equivalent ways to denote the same meaning. A corpus typically employs multiple such equivalent forms, which are identical linguistically but use different tokens, hence making them distinct entities to a computerized model. This makes it hard for the model to judge which convention to use. Normalization refers to the process of standardizing text in a document into one canonical form. It includes converting into a standard set of punctuation and character set and removing unnecessary characters which do not add to the meaning of a sentence.

For the corpus, the standard IndicNLP normalization has been used [38]. This model consists of several steps, as illustrated in Table 5 along with which languages utilize the normalization rule.

**Table 5**: Normalization rules

| Method | Languages |
|---|---|
| Convert non-standard quotation marks, hyphens and dashes to standard forms | AS, BN, EN, GU, HI, KN, ML, MR, OR, PA, TA, TE |
| Convert ellipsis to three dots | AS, BN, EN, GU, HI, KN, ML, MR, OR, PA, TA, TE |
| Remove format characters | AS, BN, EN, GU, HI, KN, ML, MR, OR, PA, TA, TE |
| Convert pipe symbol and non-standard purna virama to standard purna virama | AS, BN, GU, HI, KN, ML, MR, OR, PA, TA, TE |
| Convert colon to visarga if it follows a character in the script | AS, BN, GU, HI, KN, ML, MR, OR, TA, TE |
| Combine dependent vowels | AS, BN, KN, OR, PA, TA, TE |
| Decompose nukta-based consonants | AS, BN, HI, MR, OR |
| Convert non-standard letter forms to standard letter forms | HI, MR, OR |
| Convert old chillu encodings to corresponding new chillu encodings | ML |

The steps are elaborated as follows:

- **Converting non-standard quotation marks, hyphens, and dashes to standard forms**: Unicode consists of multiple ways of writing single and double quotation marks, hyphens, and dashes, defined under general punctuation. Apart from this there are several non-conventional ways for a person to write these punctuations as well. Therefore normalization allows to convert them to their standard forms.
- **Converting ellipsis to three dots**: Ellipsis (…) is a punctuation commonly used in verbal communication to indicate a wide variety of connotations like pause, omission, and continuation. It may also be represented by three dots (...). The IndicNLP normalizer converts the ellipsis to its three dots form.
- **Removing format characters**: Some non-printable characters are used by computerized scripts to change the format of the script. These include byte order marks (a special control character), word joiners (which remove word separations), softhyphen (a non-printable hyphen which denotes where to break a word at the end of the line, if it needs to be split among lines), and zero-width joiners and nonjoiners (used to switch between differentiated and ligated forms). These characters are not universally interpreted across systems in the same way, and do not hold any linguistic value. Hence they have been removed from the corpus.
- **Convert non-standard purna virama forms to the standard forms**: Indian languages use purna viramas (।) to denote end of the sentence. Apart from the standard purna virama, several other ways, including the pipe symbol, reserved characters, and Bengali currency numerator four, could also be used to denote purna virama. Indicnlp normalizer convert these non-standard forms to standard form.
- **Convert colon to visarga if it follows a character in the script**: Due to similarities between visarga and colon, many sources incorrectly replace colon over visarga. Thus these colons are changed to visarga. However, a colon which does not follow a character in an indic language (for example, a space), cannot be converted to a visarga.
- **Combine dependent vowels**: Some languages have a few vowel diacritics which can be written as a combination of other diacritics as well. For example, in Assamese, ে + া = ো. These diacritic combinations are replace to form the compound diacritic.
- **Decompose nukta-based consonants**: Some Indian languages have nuktas, which are attached to consonants to produce different consonants. In some cases, the consonants with nuktas are also present in the unicode set as single characters. Such consonants are replaced by the combination of the base consonants and the nuktas.
- **Convert non-standard letter forms to standard letter forms**: There are two such instances. In Marathi,    is changed to ऱ, and in Odia,    is changed to ଡ.
- **Convert old chillu encodings to new chillu encodings**: Malayalam uses chillus to denote some special consonants without vowels. In the old encodings these were represented as a combination of the base consonants and the chandrakala (the equivalent of halants in ML). However, in the new system these characters have a special place in the unicode set. Therefore these consonant-chandrakala combinations are converted into the corresponding chillu forms if they exist.

At the end, the extra whitespaces are also removed from the dataset, maintaining uniformity across text.

### 4.3.3 Concatenation and Byte Pair Encoding

The bilingual corpora set for training, validation, and testing are concatenated to form a single corpus, with a common language and all other languages on the other side. Then a byte-pair encoding (BPE) model is learned over the full dataset, using subword-nmt, with the number of workers set to 128 and the number of operations set to 48000. BPE is a form of data compression in which the most common pair of consecutive bytes are combined with a byte not present in the data [39]. The encoding model is then applied over the concatenated corpus and the full bilingual corpora set. BPE is used to deal with out-of-vocabulary (OOV) words, which are words that do not appear in the training vocabulary. These are typically uncommon or domaindependent words that aren't found in the training data but can be found in the test data. BPE performs by iteratively combining the more frequent pairs of repeated bytes or characters within the corpus, generating new tokens which represent these pairs. The process is repeated until a predetermined number of tokens has been obtained. The resultant token vocabulary is a mix of full words and subwords, where subwords are generated by the technique of merging. While training, the BPE vocabulary is used to tokenize the input sentences into subwords. This enables the model to deal with OOV words by encompassing them as a group of subwords. For instance, the word "intelligent" could be represented by the subwords "int", and "elligent".

### 4.3.4 Dictionary building and Binarization

A dictionary is built over the concatenated corpus. Given the created dictionary, the vocabulary is built over the subword-tokenized corpus, and then the corpus is binarized to make training faster and occupy less memory. The fairseq-preprocess library is used for the same. The dictionary in Fairseq is created utilizing the Byte Pair Encoding (BPE) algorithm to produce a vocabulary of subwords capable of handling rare and out-of-vocabulary words [39]. While dictionary construction, the BPE algorithm is utilized to the training data to make a vocabulary of subwords and their numerical indices. Then, the vocabulary must be saved to a file for future use while training. After constructing the dictionary, the training data is binarized to transform into a binary format which may be effectively loaded to memory during training.

## 4.4 Model Training

The experiment utilizes the Base Transformer model [40] implemented under the Fairseq library [41], an open-source sequence modeling toolkit that enables researchers to train custom models for MT tasks. The model is composed of 6 encoder-decoder layers, each with 512 hidden units and a single attention head, optimized with Adam optimizer. The dropout [42] value of 0.3 is applied to each sub-layer output before it is added to and normalized with the sub-layer input. The model is trained for a maximum of 50 epochs or 150000 updates, whichever is earlier. The model uses labelsmoothed cross entropy as the evaluation criterion. Label smoothing is an effective technique for improving model performance and is a parameter of the label-smoothed cross-entropy loss. It is done so that model is not overconfident in its prediction and acquires better ways to represent the target language. Algorithm 1 describes the steps taken to train an MNMT model.

## 4.5 Evaluation Metrics

Bilingual Evaluation Understudy (BLEU) has been used to evaluate models in the experiments. It is one of the most widely used methods for evaluating machine translation (MT) ([43]). When the hypothetical translation matches numerous strings with the reference translation, the MT evaluation gives it a higher score. The BLEU system assigns a translation a score from 0 to 1. However, it is usually represented as a percentage value(o to 100). The nearer the translation is to 100 percent, the more it corresponds to the reference translation. This matching of translation is conducted word-by-word in the same word order in both corpora. The process works by calculating the n-grams in the output by the hypothetical translation that matches the n-grams in the reference text, where 'n' represents successive words in a sentence.

$$\text{BLEU} = \text{BP} \cdot \exp\left(\sum_{n=1}^{N} w_n \log p_n\right) \quad (2)$$

In this equation (3), $p_n$ represents modified precision for $n$ gram, $w_n$ is weight in the range between 0 and 1 for $\log p_n$, such that $\sum_{n=1}^{N} w_n = 1$, and short translations face a brevity penalty (BP), described by equation (2).

$$\text{BP} = \begin{cases} 1 & \text{if } c > r \\ \exp\left(1 - \frac{r}{c}\right) & \text{if } c \leq r \end{cases} \quad (3)$$

Here, the number of unigrams in all candidate sentences is represented by $c$ whereas the best match length in the corpus for each candidate sentence is $r$. Table 3 describes the algorithm to evaluate an MNMT model.Table 6 contains scores for XX-IL, where XX is the common language. The source language is determined by the rows and the target language is determined by the columns. For example, the entry AS-BN in this table denotes the score of AS-BN from a model trained on AS-IL bilingual corpora set. Similarly, Table 7 contains scores for IL-XX. For example, the entry AS-BN in this table denotes the score of AS-BN from a model trained on IL-BN bilingual corpora set. XX-IL MNMT baseline model ranges from 0.12 to 21.35 BLEU score. AS-KN language pairs using the AS-IL model are the lowest in BLEU score and PA-HI using the PA-IL model achieves the highest BLEU score. From Table 6, it is observed that on average WI languages such as HI, GU, PA, and MR perform better than other languages as DR languages have more lexical diversity and EI languages have smaller corpus sizes. AS is extremely poor in terms of performance because of the small corpus size and biasness, since it has a significant amount of lines taken from Bible translations. TA, KN, and ML are having lower scores than the other languages. Surprisingly, TE, despite being a DR language, performs better, because of a large corpus. From Table 7, the IL-XX BLEU score ranges from 0.13 to 22.96 where HI-AS achieves the lowest score using the IL-AS model and

**Algorithm 1** Train an MNMT model

**Input**: Common language Lcom, Language set Lset, Bilingual corpora set Dset (between Lcom and each of Lset), Order O (binary: 0 indicates model from Lcom to Lset, 1 indicates model from Lset to Lcom) isTranslit (binary: 1 indicates transliteration is to be done)
**Parameters**: Number of operations in BPE S, Architecture of model Arch
Maximum number of epochs E, Stopping Criterion Crit

1: **procedure** MNMTModel(Lcom, Lset, Dset, Order, isTranslit)
2:     TokSet ← Tokenize(Dset)
3:     LowSet ← Lowercase(TokSet)
4:     NormSet ← Normalize(LowSet)
5:     RemSet ← RemoveExtraWhitespace(NormSet)
6:     **if** isTranslit = 1 **then**
7:         RemSet ← Transliterate(RemSet, Map)
8:     // Transliterate(D, M) transliterates RemSet using mapping Map (described in Algorithm XX)
9:     **end if**
10:     RemFull ← Concatenate(RemSet)
11:     BPEmodel ← BPELearn(RemFull, S)
12:     //BPELearn(D, S) learns a BPE model on corpus D performing S operations and returns the model
13:     RemFullBPE ← BPEmodel.transform(RemFull)
14:     RemSetBPE ← [BPEmodel.transform(r) for r in RemSet]
15:     DictFull ← Dictionary(RemFullBPE)
16:     BinSet ← Binarise(RemSetBPE, DictFull)
17:     **if** Order = 0 **then**
18:         LangPairs ← [(Lcom, lang) for lang in Lset]
19:     **else**
20:         LangPairs ← [(lang, Lcom) for lang in Lset]
21:     **end if**
22:     M ← TrainEarlyStopping(BinSet, LangPairs, Arch, E, Crit)
23:     // TrainEarlyStopping(D, LP, A, E, C) will train an MNMT model following architecture A on corpora set D over LP source-target language pairs for E epochs or when stopping criterion C is met (whichever is earlier) and returns the best checkpoint
24:     Return M
25. end procedure Output
26. Model M

**Algorithm 2 Evaluate an MNMT**

model

**Input**:
MNMT Model M
Source Language Lsrc
Target Language Ltgt
Bilingual test corpus Tpair (between Lsrc and Ltgt)

**Parameters**:
Evaluation Metric Met

1: **procedure** MNMTEvaluate(M, Lsrc, Ltgt, Tset)
2:     Ttr ← M.translate(Tpair[Lsrc], Lsrc, Ltgt)
3:     // M.translate(D, A, B) translates corpus D from language A to B with model M
4:     // D[A] indicates part of the corpus D in language A
5:     Score ← Met(Tpair[Ltgt], Ttr)
6: // Met(Dr, Dt) returns score in evaluation metric Met of translation Dt with respect to reference Dr
7:     Return Score
8: **end procedure**

**Output**: Evaluation metric score Score

---

PA-HI achieves the highest score using the IL-HI model. On an average languages such as AS, OR, TA, and ML accomplish the lowest score as AS, and OR is having less corpus whereas TA and ML are Dravidian languages and have high lexical diversity. HI, PA, BN, and TE achieve the highest BLEU scores as HI and PA have more corpus and are related languages.

For all languages, XX-IL in general performs better than IL-XX, with the only exceptions being AS, with insignificant differences, and HI and PA, where IL-XX performs better than XX-IL. AS has too small of a corpus, which does not give it sufficient features to learn from. Therefore XX-IL and IL-XX perform in a similar manner. The discrepancy of HI and PA can be explained via CTTR. HI and PA have the lowest CTTR, i.e. possess very less lexical diversity and contain less linguistic information. Therefore translating from HI or PA to other languages requires translating from a language with less information to a language with high information content. This leads to lower scores since sufficient information is not available for translation.

### 4.6 Language Relatedness

Grouping languages according to their families has inherent benefits as they form a group closely associated with and share numerous linguistic phenomena. A language group's vocabularies share information at the word and character levels. They have words with similar spellings that originate from the same root. In context to language families and groups, Indian languages (ILs) can be divided into two main categories:

**Table 6** XX-IL Language Baseline

| XX-IL | AS | BN | GU | HI | KN | ML | MR | OR | PA | TA | TE |
|---|---|---|---|---|---|---|---|---|---|---|---|
| AS |  | 0.48 | 0.81 | 0.97 | 0.12 | 0.16 | 0.56 | 0.31 | 0.92 | 0.52 | 0.47 |
| BN | 4.41 |  | 9.39 | 13.90 | 5.29 | 4.51 | 6.96 | 6.00 | 10.94 | 4.04 | 6.66 |
| GU | 4.52 | 7.33 |  | 18.77 | 6.79 | 4.82 | 9.34 | 7.74 | 14.3 | 4.51 | 8.67 |
| HI | 5.07 | 9.43 | 14.26 |  | 7.22 | 4.71 | 9.66 | 8.96 | 19.75 | 4.52 | 7.96 |
| KN | 3.5 | 5.07 | 7.79 | 10.63 |  | 4.00 | 5.52 | 4.49 | 8.63 | 3.4 | 7.07 |
| ML | 3.09 | 5.86 | 7.94 | 10.9 | 5.07 |  | 5.54 | 4.75 | 9 | 3.83 | 6.09 |
| MR | 3.95 | 7.27 | 10.84 | 15.12 | 5.54 | 4.73 |  | 6.69 | 12.09 | 3.74 | 6.89 |
| OR | 3.58 | 5.23 | 8.96 | 12.99 | 2.71 | 5.6 | 4.69 |  | 10.97 | 2.66 | 4.32 |
| PA | 4.27 | 6.76 | 11.24 | 21.35 | 5.52 | 3.66 | 7.1 | 7.41 |  | 3.15 | 6.28 |
| TA | 3.00 | 5.24 | 6.98 | 10.94 | 5.34 | 3.96 | 4.85 | 4.19 | 8.59 |  | 6.75 |
| TE | 3.9 | 6.82 | 9.88 | 13.58 | 7.15 | 5.93 | 6.69 | 5.64 | 10.2 | 5.17 |  |

| IL-XX | AS | BN | GU | HI | KN | ML | MR | OR | PA | TA | TE |
|---|---|---|---|---|---|---|---|---|---|---|---|
| AS | | 4.75 | 5.30 | 10.50 | 3.28 | 2.27 | 3.56 | 1.60 | 7.27 | 2.67 | 3.31 |
| BN | 0.34 | | 5.07 | 15.74 | 5.11 | 3.68 | 5.58 | 2.89 | 10.33 | 3.84 | 6.33 |
| GU | 0.33 | 6.51 | | 18.81 | 6.38 | 2.58 | 6.75 | 2.81 | 13.14 | 3.68 | 7.04 |
| HI | 0.13 | 7.24 | 11.20 | | 6.58 | 2.91 | 6.69 | 4.00 | 18.25 | 3.72 | 7.16 |
| KN | 0.27 | 4.98 | 8.06 | 14.39 | | 3.3 | 4.92 | 2.06 | 9.38 | 3.72 | 7.35 |
| ML | 0.31 | 5.64 | 6.74 | 13.6 | 5.14 | | 4.68 | 1.79 | 8.94 | 3.82 | 5.56 |
| MR | 0.36 | 7.20 | 9.31 | 16.47 | 4.99 | 3.26 | | 2.84 | 11.22 | 3.64 | 6.39 |
| OR | 0.23 | 6.04 | 8.47 | 16.09 | 5.93 | 2.83 | 5.27 | | 11.72 | 2.85 | 4.59 |
| PA | 0.2 | 6.24 | 10.23 | 22.96 | 5.51 | 3.19 | 6.52 | 3.73 | | 3.71 | 5.62 |
| TA | 0.36 | 5.06 | 6.92 | 13.17 | 7.17 | 3.12 | 3.98 | 2.06 | 8.28 | | 6.34 |
| TE | 0.34 | 6.06 | 8.70 | 15.01 | 5.14 | 3.58 | 5.34 | 2.05 | 9.65 | 4.48 | |

**Table 7** IL-XX Language Baseline

Indo-Aryan and Dravidian families. The Indo-Aryan family comprises several groups, including Eastern (Assamese, Bengali, Odia), North-western (Punjabi), Western (Gujarati), Central (Hindi), and Southern (Marathi), as illustrated in Figure 2. Owing to the degree of relatedness, the set of Indian languages is divided into three groups, Dravidian (DR) (KN, ML, TA, TE), East Indo-Aryan (EI) (AS, BN, OR), and West Indo-Aryan (WI) (GU, HI, MR, PA), where WI is just an agglomeration of language groups outside East Indo-Aryan group.

For training and evaluating, the same steps have been taken as the standard MNMT model, but the bilingual corpora set is now restricted to the language group only. For instance, while training AS-EI, only AS-BN and AS-OR bilingual corpora are considered. Tables 8, 9, and 10 show the BLEU score for related language group for
West Indo-Aryan (WI), East Indo-Aryan (EI), Dravidian (DR) languages respectively.

For WI as shown in Table 8, in the case of XX-WI, ranges lie between 8.81 to 24.27 where the PA-MR BLEU score is lowest using the PA-WI model and PA-HI scores highest using PA-WI model. For WI-XX, the BLEU score ranges between 7.76 to 23.71. PA-MR achieves the lowest using the WI-MR model and PA-HI scores the maximum BLEU score using the WI-HI model. For WI languages, related languages enhance the efficacy of the translation models owing to large corpora and low lexical diversity.

In Table 9, for XX-EI languages BLEU score lies between 0.22 to 1.01 where 0.22 is for AS-OR using the AS-EI model, and OR-BN achieves the highest BLEU score using OR-EI. AS performs worse because of less corpus size. For EI-XX, the BLEU score ranges from 0.25 to 1.24. OR-AS achieves the lowest BLEU score whereas ASBN achieves the highest BLEU score using EI-AS and AS-BN models respectively. EI in general has a negative impact on using related languages division because of less corpus.

In the case of DR Languages as shown in Table 10, for XX-DR BLEU score ranges lies from 3.49 to 8.64. ML-TA using the ML-DR model obtained a low BLEU score whereas KN-TE using the KN-DR model achieved the highest BLEU score. For DRXX, the BLEU score ranges from 2.8 to 8.13 with ML-TE as the highest BLEU score using DR-TE model whereas KN-ML is the lowest using the DR-ML model. TE performs better than the other Dravidian languages whereas ML performs the worst in both the directions. Related language does not have much impact on DR, because while on one hand these languages have sufficient corpora, whereas on the other hand, they are linguistically diverse, making their vocabularies extremely vast and difficult to generalise for a language group.

**Table 8** Related Language for West Indo-Aryan (XX-WI and WI-XX)

| XX-WI | GU    | HI    | MR    | PA    | WI-XX | GU    | HI    | MR    | PA    |
|-------|-------|-------|-------|-------|-------|-------|-------|-------|-------|
| GU    |       | 20.15 | 9.70  | 15.28 | GU    |       | 20.22 | 12.34 | 14.48 |
| HI    | 13.75 |       | 10.22 | 20.19 | HI    | 12.47 |       | 8.31  | 19.27 |
| MR    | 10.95 | 16.29 |       | 12.34 | MR    | 9.87  | 17.49 |       | 12.11 |
| PA    | 12.92 | 24.27 | 8.81  |       | PA    | 11.23 | 23.71 | 7.76  |       |

**Table 9** Related Languages for East Indo-Aryan (XX-EI and EI-XX)

| XX-EI | AS   | BN   | OR   | EI-XX | AS   | BN   | OR   |
|-------|------|------|------|-------|------|------|------|
| AS    |      | 0.66 | 0.22 | AS    |      | 1.14 | 0.61 |
| BN    | 0.71 |      | 0.99 | BN    | 0.58 |      | 0.76 |
| OR    | 0.41 | 1.01 |      | OR    | 0.25 | 0.87 |      |

**Table 10** Related Language for Dravidian Languages (XX-DR and DR-XX)

| XX-DR | KN | ML | TA | TE | DR-XX | KN | ML | TA | TE |
|---|---|---|---|---|---|---|---|---|---|
| **KN** |  | 4.66 | 4.12 | 8.64 | **KN** |  | 2.8 | 4.23 | 6.85 |
| **ML** | 4.08 |  | 3.49 | 5.18 | **ML** | 4.14 |  | 4.41 | 8.13 |
| **TA** | 5.18 | 3.94 |  | 6.98 | **TA** | 5.07 | 3.15 |  | 7.28 |
| **TE** | 5.47 | 7.51 | 5.88 |  | **TE** | 6 | 3.25 | 4.83 |  |

## 4.7 Pivot MNMT System

Pivot MT system converts among low-resourced languages using data from a high-resourcee language (Rpivot) as a "bridge" between the two [44].

In our case, translation within ILs, such as Bengali to Assamese, is typically accomplished by pivoting the English language i.e., translating Bengali (source) input to English (pivot) with a Bengali–→English (EN) model then to Assamese (target) with an English–→Assamese Model. English has been chosen as the pivot language because a lot of high-resource EN-IL bilingual corpora are available, and the EN-IL models perform very well in translation. The experiment involved using the EN-IL Samanantar corpus for the same.

Pivoting also adds an additional step of lowercasing the English language corpus. Unlike Indic languages which use one case, English uses two cases, lowercase, and uppercase. These two representations carry the same semantic meaning but are used for different grammatical contexts. For standardization and simplifying the English vocabulary, all words have been lowercased. Pivoting via EN requires developing ENIL and IL-EN models. A brief summary of the models developed is explained in the following subsection.

### 4.7.1 Analysis of EN-IL MNMT system

BLEU Scores obtained using EN-IL and IL-EN models are described in Table 11. For EN-IL, scores lie between 5.21 to 30.42 where EN-AS achieves the lowest and ENHI achieves the highest score. AS, TA, and ML obtain the lowest whereas HI, PA, and GU obtained the highest BLEU score. The former three perform poorly because AS has less corpus whereas TA and ML are agglutinative in nature and have high morphological diversity. HI, GU and BN are having qualitative as well as quantitative corpus and hence perform better.

## Table 11 EN-IL and IL-EN (BLEU scores)

| EN-IL | BLEU scores | IL-EN | BLEU scores |
|---|---|---|---|
| EN-AS | 5.21 | AS-EN | 18.64 |
| EN-BN | 17.06 | BN-EN | 28.44 |
| EN-GU | 18.63 | GU-EN | 29.28 |
| EN-HI | 30.42 | HI-EN | 33.37 |
| EN-KN | 12.13 | KN-EN | 24.69 |
| EN-ML | 9.97 | ML-EN | 26.15 |
| EN-MR | 11.66 | MR-EN | 26.90 |
| EN-OR | 10.46 | OR-EN | 24.60 |
| EN-PA | 22.15 | PA-EN | 31.03 |
| EN-TA | 7.92 | TA-EN | 23.88 |
| EN-TE | 14.49 | TE-EN | 28.62 |

In the case of IL-EN, BLEU scores lie between 18.64 to 33.37 where AS-EN is having lowest and HI-EN is having highest. For, IL-EN, AS, TA, and OR do not perform well whereas HI, PA, and GU perform better than other languages. OR does not perform well due to small corpus size. It is also noticed that IL-EN performs better than EN-IL, because ILs are lexically richer than EN.

Tables 12, 13, and 14 show the scores of related languages. In the case of EN-EI, the minimum BLEU score is achieved by EN-AS i.e. 6.9 whereas EN-BN performs the best with 17.62 BLEU score. In the case of EI-EN, AS-EN achieved the lowest 19.69 whereas OR-EN obtained the maximum BLEU score i.e. 30.84.

In the case of EN-DR, the maximum and minimum BLEU score is obtained by EN-TA i.e. 9.22 and EN-TE achieved 16.86 respectively. For DR-EN, maximum BLEU scores are obtained by TA-EN i.e. 23.85, and lowest scores are obtained by TE-EN i.e. 28.56.

In the case of EN-WI, scores lie between 13.2 obtained by EN-MR and 32.54 obtained by EN-HI. WI-EN lies between 27.37 to 34.33 where MR-EN achieves the lowest and HI-EN achieves the highest scores.

As shown in Tables 11, 12, 13, and 14, the use of related languages benefits the development of EN-IL and IL-EN models, observed via an increase in scores.

## Table 12 Related Language Scores for West Indo-Aryan (EN-WI and WI-EN)

| Languages | BLEU | Languages | BLEU |
|---|---|---|---|
| EN-GU | 19.41 | GU-EN | 30.79 |
| EN-HI | 32.54 | HI-EN | 34.33 |
| EN-MR | 13.20 | MR-EN | 27.37 |
| EN-PA | 23.32 | PA-EN | 32.02 |

**Table 13** Related Language Scores for East Indo-Aryan (EN-EI and EI-EN)

| Languages | BLEU | Languages | BLEU |
|---|---|---|---|
| EN-AS | 6.90 | AS-EN | 19.69 |
| EN-BN | 17.62 | BN-EN | 30.84 |
| EN-OR | 11.54 | OR-EN | 25.44 |

**Table 14** Related Language for Dravidian Languages (EN-DR and DR-EN)

| Languages | BLEU | Languages | BLEU |
|---|---|---|---|
| EN-KN | 13.78 | KN-EN | 24.24 |
| EN-ML | 11.35 | ML-EN | 26.86 |
| EN-TA | 9.22 | TA-EN | 23.85 |
| EN-TE | 16.86 | TE-EN | 28.56 |

#### 4.7.2 Development and Training of Pivot MNMT System

Training a pivot MNMT system involves training two models, one from source to pivot and one from pivot to target. As it has been established that the use of related languages benefits EN-IL models, henceforth the related language models are used to develop pivot models. For instance, developing a GU-EN-KN model requires training in WI-EN and EN-DR models, since GU falls in the WI group and KN falls in the DR group. After this, the GU text is translated to EN using WI-EN, and the EN translation is translated to KN using EN-DR. The algorithm for the development of the pivot model is explained in Algorithm 3 and Algorithm 4 describes the process to train and evaluate a pivot MNMT model respectively.

**Algorithm 3** Train a pivot MNMT model

**Input**:
Common language Lcom
Language sets LsetSrc and LsetTgt (Source and Target side respectively) Bilingual corpora sets DsetSrc and DsetTgt (Source and Target side respectively) is Translit (binary: 1 indicates transliteration is to be done)

**Parameters**:
Number of operations in BPE S
Architecture of model Arch
Maximum number of epochs E
Stopping Criterion Crit

1: **procedure** PivotMNMTModel(Lcom, LsetSrc, LsetTgt, DsetSrc, DsetTgt, isTranslit)
2:     Msrc ← MNMTModel(Lcom, LsetSrc, DsetSrc, 1, isTranslit)
3:     Mtgt ← MNMTModel(Lcom, LsetTgt, DsetTgt, 0, isTranslit)
4:     Return Msrc, Mtgt
5: **end procedure**

**Output**: Source-side model Msrc and Target-side model Mtgt

---

For the IL-IL pivot MNMT model, EN-DR, EN-EI, and EN-WI are utilized. Table 15 shows the result of the Pivot Language Baseline for IL-IL. From the result, it is observed that AS-TA achieved a minimum BLEU score (4.65 BLEU) using EI-ENDR and PA-HI achieved the highest score (23.24 BLEU) using WI-EN-WI. AS, ML, and TA achieve the lowest BLEU score whereas HI, PA, and GU perform better.

The best scores with and without pivot are illustrated in Table 16.

Tables 15 and 16 show that pivot scores perform better than related IL-IL, with some exceptions. The exceptions are described in Table 17.

### 4.8 Transliteration

Although ILs have several similarities, most languages use distinct scripts. The use of distinct scripts may cause redundancy in dictionary formation where the same token is recorded twice in different scripts. Converting all scripts to the same script improves shared vocabulary and reduces subword vocabulary [33].

---

**Algorithm 4** Evaluate a pivot MNMT model

**Input**:
Pivot MNMT Models Msrc, Mtgt
Source Language Lsrc
Target Language Ltgt
Pivot Language Lpvt
Bilingual test corpus Tpair (between Lsrc and Ltgt)

**Parameters**:
Evaluation Metric Met

1: **procedure** MNMTEvaluate(Msrc, Mtgt, Lsrc, Ltgt, Lpvt, Tset)
2:     Tpt ← Msrc.translate(Tpair[Lsrc], Lsrc, Lpvt)
3:     Ttr ← Mtgt.translate(Tpt, Lpvt, Ltgt)
4:     Score ← Met(Tpair[Ltgt], Ttr)

        5:    Return Score
6: **end procedure**

**Output**: Evaluation metric score Score

**Table 15 Using Pivot Language Baseline**

| Pivot | AS | BN | GU | HI | KN | ML | MR | OR | PA | TA | TE |
|---|---|---|---|---|---|---|---|---|---|---|---|
| AS |  | 8.42 | 9.61 | 14.1 | 5.89 | 4.99 | 6.99 | 6.07 | 11.18 | 4.65 | 7.63 |
| BN | 6.49 |  | 13.13 | 19.73 | 8.71 | 7.43 | 9.31 | 8.79 | 15.42 | 5.96 | 10.96 |
| GU | 6.31 | 11.93 |  | 20.8 | 9.06 | 7.98 | 10.08 | 8.3 | 16.08 | 6.53 | 10.85 |
| HI | 5.92 | 11.76 | 13.62 |  | 9.43 | 7.43 | 9.74 | 8.17 | 17.48 | 6.05 | 9.98 |
| KN | 4.69 | 9.61 | 6.67 | 5.45 |  | 9.5 | 10.87 | 6.78 | 16.84 | 8.06 | 13.25 |
| ML | 5.06 | 10.87 | 11.42 | 8.26 | 17.92 |  | 8.05 | 6.85 | 13.87 | 5.88 | 10.03 |
| MR | 5.48 | 11.57 | 12.63 | 19.5 | 8.26 | 6.72 |  | 7.52 | 15.11 | 5.67 | 9.9 |
| OR | 5.73 | 10.87 | 12.14 | 18.49 | 8.19 | 6.72 | 9.08 |  | 14.47 | 5.21 | 9.42 |
| PA | 5.94 | 12.20 | 13.7 | 23.24 | 9.04 | 7.69 | 9.63 | 8.36 |  | 5.79 | 10.64 |
| TA | 4.98 | 9.89 | 10.87 | 16.42 | 8.24 | 6.67 | 7.8 | 6.83 | 13.01 |  | 9.36 |
| TE | 5.55 | 10.78 | 12.42 | 18.51 | 8.84 | 7.53 | 9.05 | 7.44 | 13.81 | 6.19 |  |

For the purposes of the experiment, the ITRANS encoding has been utilized for transliteration. ITRANS is an ASCII transliteration scheme for ILs, specifically Devanagari script. It is preferred over other schemes because it has a wide scope of mapping, and uses only standard English characters without the use of accents. However, ITRANS also provides a transliteration scheme for other languages, but with several shortcomings, such as ambiguous or no mappings for Dravidian vowels, nuktas, and chillus. Therefore a modified version of ITRANS has been utilized. The mapping is illustrated in Table 18.

**Table 16** Best scores with and without pivot

| Best Scores | AS | BN | GU | HI | KN | ML | MR | OR | PA | TA | TE |
|---|---|---|---|---|---|---|---|---|---|---|---|
| AS |  | 8.42 | 9.61 | 14.1 | 5.89 | 4.99 | 6.99 | 6.07 | 11.18 | 4.65 | 7.63 |
| BN | 6.49 |  | 13.13 | 19.73 | 8.71 | 7.43 | 9.31 | 8.79 | 15.42 | 5.96 | 10.96 |
| GU | 6.31 | 11.93 |  | 20.8 | 9.06 | 7.98 | 10.08 | 8.3 | 16.08 | 6.53 | 10.85 |
| HI | 5.92 | 11.76 | 14.26 |  | 9.43 | 7.43 | 9.74 | 8.96 | 19.75 | 6.05 | 9.98 |
| KN | 4.69 | 9.61 | 8.06 | 14.39 |  | 9.5 | 10.87 | 6.78 | 16.84 | 8.06 | 13.25 |
| ML | 5.06 | 10.87 | 11.42 | 8.26 | 17.92 |  | 8.05 | 6.85 | 13.87 | 5.88 | 10.03 |
| MR | 5.48 | 11.57 | 12.63 | 19.5 | 8.26 | 6.72 |  | 7.52 | 15.11 | 5.67 | 9.9 |
| OR | 5.73 | 10.87 | 12.14 | 18.49 | 8.19 | 6.72 | 9.08 |  | 14.47 | 5.21 | 9.42 |
| PA | 5.94 | 12.2 | 13.7 | 23.24 | 9.04 | 7.69 | 9.63 | 8.36 |  | 5.79 | 10.64 |
| TA | 4.98 | 9.89 | 10.87 | 16.42 | 8.24 | 6.67 | 7.8 | 6.83 | 13.01 |  | 9.36 |
| TE | 5.55 | 10.78 | 12.42 | 18.51 | 8.84 | 7.53 | 9.05 | 7.44 | 13.81 | 6.19 |  |

## Table 17 Cases where pivot decreases scores

| Language Pair | Pivot Score | Best Model without Pivot | Best Score |
|---|---:|---|---:|
| **GU-MR** | 10.08 | WI-MR | 12.34 |
| **HI-GU** | 13.62 | HI-IL | 14.26 |
| **HI-MR** | 9.74 | HI-WI | 10.22 |
| **HI-OR** | 8.17 | HI-IL | 8.96 |
| **HI-PA** | 17.48 | HI-WI | 20.19 |
| **KN-GU** | 6.67 | IL-GU | 8.06 |
| **KN-HI** | 5.45 | IL-HI | 14.39 |
| **PA-HI** | 23.24 | PA-WI | 24.27 |

Algorithm 5 describes the process of transliteration.

IndicTrans has been used in the experiment for transforming all Indic Language data into ITRANS by mapping distinct Unicode ranges. Among others, IndicTrans offers both itrans and itrans_dravidian schemes, where the latter provides support for long Dravidian vowels 'E' and 'O'. After that, some postprocessing techniques are also applied to make the mapping more accurate, as shown in the table.

However, since ITRANS is a Roman script, and the Samanantar Indic corpus contains noisy English data, it will be difficult for the computer to distinguish between this noise and actual Indic words after transliteration. Therefore, before transliteration, English letters are first mapped to characters of a foreign language (in this case Thai (TH)). These TH embeddings stay unaffected during transliteration and thus indicate the presence of English in the system. The embeddings are obtained by computing the integer value of the English characters, adding 0x0DC0, and changing it back to the corresponding character.

---

**Algorithm 5** Transliterate a bilingual corpora set

---

**Input**:
Bilingual corpora set Dset
Map M

**Assumption**: Every character in the map M in source script is mapped to exactly one character in the target script

1: **procedure** Transliterate(Dset, M)
2:     Let TlSet be an empty bilingual corpora set
3:     **for each** corpus Dpair ϵDset **do**
4:         Let TlPair be an empty bilingual corpus

```
5:         TlPair.languages ← DPair.languages
6:             // Dpair.languages returns language pair of bilingual corpus Dpair
7:         for each language L ∈TlPair.languages do
8:             for each line Ln ∈Dpair[L] do
9:  Let NewLn be an empty string
10:                 for each character Ch ∈Ln do
11:                     if Ch is an English letter then
12:                         NewLn.append(chr(ord(Ch) + 0x0DC0))
13:                         // A.append(B) adds B to the end of A
14:                         // ord(A) returns integer value of character A
15:                         // chr(A) returns character with integer value A
16:                     else
17:                         NewLn.append(M[Ch])
18:  // M[C] returns the corresponding mapping of character C in map M
19:                     end if
20:                 end for
21:                 TlPair[L].append(NewLn)
22:             end for
23:         end for
24:         TlSet.append(TlPair)
25:     end for
26:     Return TlSet
27: end procedure
```

**Output**: Transliterated bilingual corpora set TlSet

---

For building the transliteration models, the best model is considered from the previous stages and a model is trained on the corresponding transliterated corpus. In most cases, it involves training a pivot model using related groups, except the cases mentioned in Table 17, where the better model is used. Tables 19, 20, and 21 depict the scores of models on transliterated EN-WI, EN-EI and EN-DR in both directions.

Table 19 Transliteration score for West Indo-Aryan (EN-WI and WI-EN)

| Languages | BLEU | Languages | BLEU |
|---|---|---|---|
| **EN-GU** | 20.56 | GU-EN | 31.43 |
| **EN-HI** | 32.53 | HI-EN | 34.86 |
| **EN-MR** | 16.06 | MR-EN | 28.58 |
| **EN-PA** | 24.18 | PA-EN | 32.73 |

Table 20 Transliteration Language score for East Indo-Aryan (EN-EI and EI-EN)

| Languages | BLEU | Languages | BLEU |
|---|---|---|---|
| **EN-AS** | 9.89 | AS-EN | 20.79 |
| **EN-BN** | 20.27 | BN-EN | 30.85 |
| **EN-OR** | 16.03 | OR-EN | 25.59 |
| **EN-PA** | 24.18 | PA-EN | 32.73 |

Table 21 Transliteration Language for Dravidian Language (EN-DR and DR-EN)

| Languages | BLEU | Languages | BLEU |
|---|---|---|---|
| **EN-KN** | 14.28 | KN-EN | 25.45 |
| **EN-ML** | 22.31 | ML-EN | 27.54 |
| **EN-TA** | 15.24 | TA-EN | 25.26 |
| **EN-TE** | 19.75 | TE-EN | 29.46 |

The tables show that transliteration causes significant increments in the model performance. This is because, without transliteration, the system treated the same token written in different languages as different entities, which prevented it from making a truly shared vocabulary. However, transliterating into the same script allows these tokens to now be treated as the same, thus promoting language features to be learned across different languages. This also explains why models with lower scores benefit more from transliteration than the better ones.

In the case of EN-WI, the scores range from 16.06 to 32.53 where MR and HI achieve the highest and lowest scores respectively. For HI there is no increment whereas for PA, GU, and MR there is an increment of about 1, 1, and 3 BLEU on average. In the case of WI-EN, the scores lie between 28.56, achieved by MR-EN, and 34.86, achieved by HI-EN. All languages oversee an increase of about 1 BLEU on average.

For EN-EI the BLEU score ranges between 9.89 to 20.27 where EN-AS and EN-BN achieve the lowest and the highest scores respectively. AS and BN show an increase of about 3 and OR shows an increase of about 5 BLEU scores on average. For EI-EN the BLEU scores range from 20.79 in AS-EN to 30.85 in BN-EN where AS and BN show an increment of about 1 BLEU while OR does not show significant changes on average.

In the case of EN-DR, the BLEU score lies between 14.28 to 22.31 whereas ENKN and EN-ML is having the lowest and highest BLEU scores. It is observed that ML, KN, TA, and TE have an increase in BLEU scores of 11, 5, 6, and 3 respectively. For DR-EN, the BLEU score lies between 25.26 to 29.46 where TA and TE achieve maximum and minimum scores respectively, where all languages show an increase of about 1 BLEU on average.

Table 22 shows the scores obtained by using transliteration. Table 23 shows the language pairs which are the exception case and transliteration does not perform better.

The BLEU score of transliteration models ranges from 7.03 to 24.29 where minimum scores are obtained in AS-KN using the EI-EN-DR model whereas maximum scores are achieved by PA-HI using the PA-WI model. The maximum BLEU score is obtained by HI, PA, and BN whereas AS, KN, and OR achieve the minimum BLEU score on average.

In general, from all the models, the increment of the score is observed in ML, TA, and BN has an increment of 6.67, 4.72, and 4.24 BLEU scores. The worst average increment is observed in KN, HI, and PA with BLEU scores of 1.45, 1.83, and 2.67.

**Table 23 Evaluation Metrics of exception pair with and without Transliteration**

| Language Pair | Model | Without translit | With translit |
|---|---|---|---|
| **GU-MR** | WI-MR | 12.34 | 12.14 |
| **HI-GU** | HI-IL | 14.26 | 13.77 |
| **KN-PA** | DR-EN-WI | 16.84 | 14.67 |
| **KN-TE** | DR-EN-DR | 13.25 | 11.11 |
| **OR-KN** | EI-EN-DR | 8.19 | 8.09 |
| **KN-HI** | IL-HI | 14.39 | 14.14 |

**Table 22 Transliteration Score**

| Translit Scores | AS | BN | GU | HI | KN | ML | MR | OR | PA | TA | TE |
|---|---|---|---|---|---|---|---|---|---|---|---|
| AS |  | 13.07 | 12.6 | 14.94 | 7.03 | 15.04 | 10.81 | 9.97 | 12.73 | 10.49 | 11.76 |
| BN | 10.98 |  | 16.58 | 20.19 | 10.36 | 18.14 | 14.14 | 13.11 | 17.03 | 13.2 | 14.88 |
| GU | 10.25 | 17.26 |  | 21.7 | 10.46 | 18.56 | 12.14 | 12.98 | 17.74 | 13.6 | 15 |
| HI | 8.58 | 14.91 | 13.77 |  | 9.84 | 17.23 | 11.23 | 12.56 | 20.72 | 10.54 | 12.44 |
| KN | 7.46 | 13.2 | 8.5 | 14.14 |  | 15.78 | 10.41 | 11.17 | 14.67 | 10.12 | 11.11 |
| ML | 8.76 | 15.27 | 14.36 | 18.96 | 9.49 |  | 12.81 | 11.43 | 15.25 | 12.68 | 14.16 |
| MR | 9.62 | 16.43 | 15.62 | 20.37 | 9.85 | 17.94 |  | 12.46 | 16.43 | 13.02 | 14.65 |
| OR | 7.74 | 13.76 | 13.47 | 18.85 | 8.09 | 15.44 | 11.22 |  | 15.59 | 9.17 | 11.07 |
| PA | 9.22 | 16.09 | 16.6 | 24.29 | 9.84 | 18.35 | 13.38 | 12.81 |  | 12.05 | 13.88 |
| TA | 8.36 | 13.99 | 13.62 | 17.7 | 9.31 | 17.32 | 11.75 | 10.93 | 14.22 |  | 13.13 |
| TE | 9.8 | 15.98 | 15.41 | 19.44 | 9.5 | 18.63 | 13.28 | 11.99 | 16.05 | 13.27 |  |

## 4.9 Best Scores

From all the models, the best scores that are obtained are shown are described in Table 24. The BLEU score of Best score models ranges from 7.03 to 24.29 where minimum scores are obtained in AS-KN using the EI-EN-DR model (both transliteration models) whereas maximum scores are achieved by PA-HI using the PA-WI transliteration model. The maximum BLEU score is obtained by HI, PA, and BN whereas AS, KN, and OR achieve the minimum BLEU score on average.

**Table 24 Final best scores of all approaches**

| Final Scores | AS | BN | GU | HI | KN | ML | MR | OR | PA | TA | TE |
|---|---|---|---|---|---|---|---|---|---|---|---|
| AS |  | 13.07 | 12.6 | 14.94 | 7.03 | 15.04 | 10.81 | 9.97 | 12.73 | 10.49 | 11.76 |
| BN | 10.98 |  | 16.58 | 20.19 | 10.36 | 18.14 | 14.14 | 13.11 | 17.03 | 13.2 | 14.88 |
| GU | 10.25 | 17.26 |  | 21.7 | 10.46 | 18.56 | 12.34 | 12.98 | 17.74 | 13.6 | 15 |
| HI | 8.58 | 14.91 | 14.26 |  | 9.84 | 17.23 | 11.23 | 12.56 | 20.72 | 10.54 | 12.44 |
| KN | 7.46 | 13.2 | 8.5 | 14.39 |  | 15.78 | 10.41 | 11.17 | 16.84 | 10.12 | 13.25 |
| ML | 8.76 | 15.27 | 14.36 | 18.96 | 9.49 |  | 12.81 | 11.43 | 15.25 | 12.68 | 14.16 |
| MR | 9.62 | 16.43 | 15.62 | 20.37 | 9.85 | 17.94 |  | 12.46 | 16.43 | 13.02 | 14.65 |
| OR | 7.74 | 13.76 | 13.47 | 18.85 | 8.19 | 15.44 | 11.22 |  | 15.59 | 9.17 | 11.07 |
| PA | 9.22 | 16.09 | 16.6 | 24.29 | 9.84 | 18.35 | 13.38 | 12.81 |  | 12.05 | 13.88 |
| TA | 8.36 | 13.99 | 13.62 | 17.7 | 9.31 | 17.32 | 11.75 | 10.93 | 14.22 |  | 13.13 |
| TE | 9.8 | 15.98 | 15.41 | 19.44 | 9.5 | 18.63 | 13.28 | 11.99 | 16.05 | 13.27 |  |

Table 18: Transliteration mapping

| Map | AS | BN | GU | HI/MR | KN | ML | OR | PA | TA | TE |
|---|---|---|---|---|---|---|---|---|---|---|
| a | অ | অ | અ | अ | ಅ | അ | ଅ | ਅ | அ | అ |
| a[1] | | | | | | | | ਅੱ | | |
| A | আ | আ | આ | आ | ಆ | ആ | ଆ | ਆ | ஆ | ఆ |
| i | ই | ই | ઇ | इ | ಇ | ഇ | ଇ | ਇ | இ | ఇ |
| I | ঈ | ঈ | ઈ | ई | ಈ | ഈ | ଈ | ਈ | ஈ | ఈ |
| u | উ | উ | ઉ | उ | ಉ | ഉ | ଉ | ਉ | உ | ఉ |
| U | ঊ | ঊ | ઊ | ऊ | ಊ | ഊ | ଊ | ਊ | ஊ | ఊ |
| RRi | ঋ | ঋ | ઋ | ऋ | ಋ | ഋ | ଋ | | | ఋ |
| e | এ | এ | એ | ए | ಎ | എ | ଏ | ਏ | எ | ఎ |
| E | | | | | ಏ | ഏ | | | ஏ | ఏ |
| ai | ঐ | ঐ | ઐ | ऐ | ಐ | ഐ | ଐ | ਐ | ஐ | ఐ |
| o | ও | ও | ઓ | ओ | ಒ | ഒ | ଓ | ਓ | ஒ | ఒ |
| O | | | | | ಓ | ഓ | | | ஓ | ఓ |
| au | ঔ | ঔ | ઔ | औ | ಔ | ഔ | ଔ | ਔ | ஔ | ఔ |
| M | অং | অং | અં | अं | ಅಂ | അം | ଅଂ | ਅਂ/ਅੰ | அஂ | అం |
| H | অঃ | অঃ | અઃ | अः | ಅಃ | അഃ | ଅଃ | | அஃ | అః |

[1] does not occur at the end of a word. Doubles the consonant sound after it (eg- ਅੱਜ: ajja)

Table 18: Transliteration mapping

| Map | AS | BN | GU | HI/MR | KN | ML | OR | PA | TA | TE |
|---|---|---|---|---|---|---|---|---|---|---|
| m | ম্ | ম্ | મ્ | म् | ಮ್ | മ് | ମ୍ | ਮ੍ | ம் | మ్ |
| y | য্ | য্ | ય્ | य् | ಯ್ | യ് | ଯ୍ | ਯ੍ | ய் | య్ |
| Y | য়্ | য়্ | ય્ | य़् |  |  | ୟ୍ | ਯ਼੍ |  |  |
| r | ৰ্ | র্ | ર્ | र् | ರ್ | ര് | ର୍ | ਰ੍ | ர் | ర్ |
| R |  |  | ર્ | ऱ् | ಱ್ | റ് |  |  | ற் | ఱ్ |
| ಱೆ |  |  |  |  |  | റ് |  |  |  |  |
| l | ল্ | ল্ | લ્ | ल् | ಲ್ | ല് | ଲ୍ | ਲ੍ | ல் | ల్ |
| ಳೆ |  |  |  |  |  | ള് |  |  |  |  |
| L |  |  | ળ્ | ळ् | ಳ್ | ള് | ଳ | ਲ਼੍ | ள் | ళ్ |
| ಳೆ |  |  |  |  |  | ഌ് |  |  |  |  |
| zh |  |  |  |  |  | ഴ് |  | ਯ਼ |  |  |
| v | ৱ্ | ব্ | વ્ | व् | ವ್ | വ് | ୱ | ਦ੍ | வ் | వ్ |
| w |  |  |  | व़् |  |  |  |  |  |  |
| sh | শ্ | শ্ | શ્ | श् | ಶ್ | ശ് | ଶ୍ | ਸ਼੍ | ஶ் | శ్ |
| Sh | ষ্ | ষ্ | ષ્ | ष् | ಷ್ | ഷ് | ଷ |  | ஷ் | ష్ |
| s | স্ | স্ | સ્ | स् | ಸ್ | സ് | ସ | ਸ੍ | ஸ் | స్ |
| h | হ্ | হ্ | હ્ | ह् | ಹ್ | ഹ് | ହ | ਹ੍ | ஹ் | హ్ |
| 0 | ০ | ০ | ૦ | ० | ೦ | ൦ | ୦ | ੦ | ௦ | ౦ |
| 1 | ১ | ১ | ૧ | १ | ೧ | ൧ | ୧ | ੧ | ௧ | ౧ |
| 2 | ২ | ২ | ૨ | २ | ೨ | ൨ | ୨ | ੨ | ௨ | ౨ |
| 3 | ৩ | ৩ | ૩ | ३ | ೩ | ൩ | ୩ | ੩ | ௩ | ౩ |
| 4 | ৪ | ৪ | ૪ | ४ | ೪ | ൪ | ୪ | ੪ | ௪ | ౪ |
| 5 | ৫ | ৫ | ૫ | ५ | ೫ | ൫ | ୫ | ੫ | ௫ | ౫ |
| 6 | ৬ | ৬ | ૬ | ६ | ೬ | ൬ | ୬ | ੬ | ௬ | ౬ |
| 7 | ৭ | ৭ | ૭ | ७ | ೭ | ൭ | ୭ | ੭ | ௭ | ౭ |
| 8 | ৮ | ৮ | ૮ | ८ | ೮ | ൮ | ୮ | ੮ | ௮ | ౮ |
| 9 | ৯ | ৯ | ૯ | ९ | ೯ | ൯ | ୯ | ੯ | ௯ | ౯ |

Table 18: Transliteration mapping

| Map | AS | BN | GU | HI/MR | KN | ML | OR | PA | TA | TE |
|---|---|---|---|---|---|---|---|---|---|---|
| .N | অঁ | অঁ | અં | अँ |  |  | ଅଁ |  |  | అఁ |
| aI |  |  |  | ऑ |  |  |  |  |  |  |
| k | ক্ | ক্ | ક્ | क् | ಕ್ | ക് | କ | ਕ੍ |  | క్ |
| q |  |  | ક્ | क़् |  |  | କ. | ਕ੍ |  |  |
| kh | খ্ | খ্ | ખ્ | ख् | ಖ್ | ഖ് | ଖ | ਖ੍ |  | ఖ్ |
| K |  |  | ખ્ | ख़् |  |  | ଖ. | ਖ੍ |  |  |
| g | গ্ | গ্ | ગ્ | ग् | ಗ್ | ഗ് | ଗ | ਗ੍ |  | గ్ |
| G |  |  | ગ્ | ग़् |  |  | ଗ. | ਗ੍ |  |  |
| gh | ঘ্ | ঘ্ | ઘ્ | घ् | ಘ್ | ഘ് | ଘ | ਘ੍ | ఁ | ఘ్ |
| ~N | ঙ্ | ঙ্ | ઙ્ | ङ् | ಙ್ |  | ଙ | ਙ੍ | ங | ఙ్ |
| ch | চ্ | চ্ | ચ્ | च् | ಚ್ | ച് | ଚ | ਚ੍ |  | చ్ |
| cha |  |  |  |  |  |  | ଚ. |  |  |  |
| Ch | ছ্ | ছ্ | છ્ | छ् | ಛ್ | ഛ് | ଛ | ਛ੍ |  | ఛ్ |
| j | জ্ | জ্ | જ્ | ज् | ಜ್ | ജ് | ଜ | ਜ੍ | ஜ | జ్ |
| z |  |  | જ઼્ | ज़् |  |  | ଜ. |  |  |  |
| J |  |  |  |  |  |  |  | ਜ੍ |  |  |
| jh | ঝ্ | ঝ্ | ઝ્ | झ् | ಝ್ | ഝ് | ଝ | ਝ੍ | ஂ | ఝ్ |
| ~n | ঞ্ | ঞ্ | ઞ્ | ञ् | ಞ್ | ഞ് | ଞ | ਞ੍ | ஞ | ఞ్ |
| T | ট্ | ট্ | ટ્ | ट् | ಟ್ | ട് | ଟ | ਟ੍ |  | ట్ |
| Th | ঠ্ | ঠ্ | ઠ્ | ठ् | ಠ್ | ഠ് | ଠ | ਠ੍ |  | ఠ్ |
| D | ড্ | ড্ | ડ્ | ड् | ಡ್ | ഡ് | ଡ | ਡ੍ |  | డ్ |
| .D |  |  | ડ઼્ | ड़् |  |  | ଡ଼ | ਡ੍ |  |  |
| Dh | ঢ্ | ঢ্ | ઢ્ | ढ् | ಢ್ | ഢ് | ଢ | ਢ੍ | ఁ | ఢ్ |
| .Dh |  |  | ઢ઼્ | ढ़् |  |  | ଢ଼ | ਢ੍ |  |  |
| N | ণ্ | ণ্ | ણ્ | ण् | ಣ್ | ണ് | ଣ | ਣ੍ | ண | ణ్ |
| ണ |  |  |  |  |  | ണ |  |  |  |  |
| t | ত্ | ত্ | ત્ | त् | ತ್ | ത് | ତ | ਤ੍ |  | త్ |
| t[2] | ৎ | ৎ |  |  |  |  |  |  |  |  |
| th | থ্ | থ্ | થ્ | थ् | ಥ್ | ഥ് | ଥ | ਥ੍ |  | థ్ |
| d | দ্ | দ্ | દ્ | द् | ದ್ | ദ് | ଦ | ਦ੍ |  | ద్ |
| dh | ধ্ | ধ্ | ધ્ | ध् | ಧ್ | ധ് | ଧ | ਧ੍ | ఁ | ధ్ |
| n | ন্ | ন্ | ન્ | न् | ನ್ | ന് | ନ | ਨ੍ | ந | న్ |
| ന |  |  |  |  |  | ന |  |  |  |  |
| na |  |  |  | ऩ् |  |  |  |  |  |  |
| n2 |  |  |  |  | ನ್ |  |  |  | ன |  |
| p | প্ | প্ | પ્ | प् | ಪ್ | പ് | ପ | ਪ੍ |  | ప్ |
| ph | ফ্ | ফ্ | ફ્ | फ् | ಫ್ | ഫ് | ଫ | ਫ੍ |  | ఫ్ |
| f |  |  | ફ઼્ | फ़् |  |  | ଫ. | ਫ੍ |  |  |
| b | ব্ | ব্ | બ્ | ब् | ಬ್ | ബ് | ବ | ਬ੍ |  | బ్ |
| bh | ভ্ | ভ্ | ભ્ | भ् | ಭ್ | ഭ് | ଭ | ਭ੍ | ఁ | భ్ |

[2]only occurs at the end of a word

## 5  Conclusion and Future Work

The work involves the lexical diversity of the Samanantar corpus, which revealed that HI and PA are lexically poor and TA and ML are lexically rich among all ILs. Baselines for IL-IL are presented using the MNMT model with and without using related languages. For the development of the pivot models, EN-IL models are similarly built. It is observed in the case of IL-IL models that the concept of related languages helps the WI group only but adversely affects the EI group. Nonetheless, the concept increases the performance of EN-IL models. It is also noticed that using English as a pivot and transliterating into the proposed modified ITRANS script increases the scores of the models with few exceptions. On the one hand, using related languages is helpful for relatively high-resource languages, whereas on the other hand using pivot and transliteration are beneficial for relatively low-resource languages. Moreover, transliteration significantly improves models built for lexically rich languages like ML and TA. In future work, the Zero-short translation method will be explored. Zero-shot translation refers to an MNMTT model that can perform translations among language pairs for which it has not been specifically trained.

## 6  Acknowledgement

This work is partially funded by Meity (Ministry of Electronics and Information Technology, Government of India) for project sanction no. 13 (12)/2020-CC & BT dated 24.04.2020.